\pdfoutput=1

\documentclass[11pt]{article}

\usepackage[final]{acl}

\usepackage{times}
\usepackage{latexsym}

\usepackage[T1]{fontenc}

\usepackage[utf8]{inputenc}

\usepackage{microtype}

\usepackage{inconsolata}
\usepackage{amsfonts}
\usepackage{amsmath}
\usepackage{array}
\usepackage{natbib}
\usepackage{float}
\usepackage{graphicx}
\usepackage{subcaption}
\usepackage{tcolorbox}
\usepackage{hyperref}
\usepackage{multirow}
\usepackage{multicol}
\usepackage{makecell}
\usepackage{booktabs}
\usepackage{enumitem}
\usepackage{stfloats}
\usepackage{pgfplots}
\usepackage{pgfplotstable}
\usepackage{adjustbox}
\pgfplotsset{compat=newest}
\usepgfplotslibrary{colorbrewer}
\usepackage{siunitx,xcolor,colortbl}
\tcbuselibrary{breakable} 

\definecolor{mygray}{rgb}{0.95,0.95,0.95} 
\definecolor{darkgreen}{RGB}{45, 130, 117}
\definecolor{darkorange}{RGB}{187, 114, 14}
\definecolor{myblue}{RGB}{65, 85, 198}
\definecolor{darkred}{RGB}{192, 0, 0}
\newcommand{\opadd}{$+$}
\newcommand{\opsub}{$-$}
\newcommand{\opmul}{$*$}
\newcommand{\opdiv}{$/$}
\title{Generator-Assistant Stepwise Rollback Framework \\for Large Language Model Agent}

\author{Xingzuo Li\textsuperscript{\rm $1$}, Kehai Chen\textsuperscript{\rm $1$}, Yunfei Long\textsuperscript{\rm $2$}, Xuefeng Bai\textsuperscript{\rm $1$}\thanks{Corresponding author.}, Yong Xu\textsuperscript{\rm $1$}, Min Zhang\textsuperscript{\rm $1$} \\
$\textsuperscript{\rm $1$}$School of Computer Science and Technology, Harbin Institute of Technology, Shenzhen, China \\
$\textsuperscript{\rm $2$}$School of Electronic Engineering and Computer Science, Queen Mary University of London \\
\texttt{24S051028@stu.hit.edu.cn}, \texttt{qp241311@qmul.ac.uk},\\
\texttt{\{chenkehai, baixuefeng, laterfall, zhangmin2021\}@hit.edu.cn}
}

\begin{document}
\maketitle
\begin{abstract}

Large language model (LLM) agents typically adopt a step-by-step reasoning framework, in which they interleave the processes of thinking and acting to accomplish the given task.
However, this paradigm faces a deep-rooted \textit{one-pass} issue whereby each generated intermediate thought is plugged into the trajectory regardless of its correctness, which can cause irreversible error propagation.
To address the issue, this paper proposes a novel framework called Generator-Assistant Stepwise Rollback (GA-Rollback) to induce better decision-making for LLM agents. 
Particularly, GA-Rollback utilizes a generator to interact with the environment and an assistant to examine each action produced by the generator, where the assistant triggers a rollback operation upon detection of incorrect actions.
Moreover, we introduce two additional strategies tailored for the rollback scenario to further improve its effectiveness. 
Extensive experiments show that GA-Rollback achieves significant improvements over several strong baselines on three widely used benchmarks. 
Our analysis further reveals that GA-Rollback can function as a robust plug-and-play module, integrating seamlessly with other methods. \footnote{
Our code is available at \url{https://github.com/wisper12933/GA-Rollback}.
} 

\end{abstract}

\section{Introduction}

\begin{figure}[pt]
    \centering
    {
    \includegraphics[width=\linewidth]{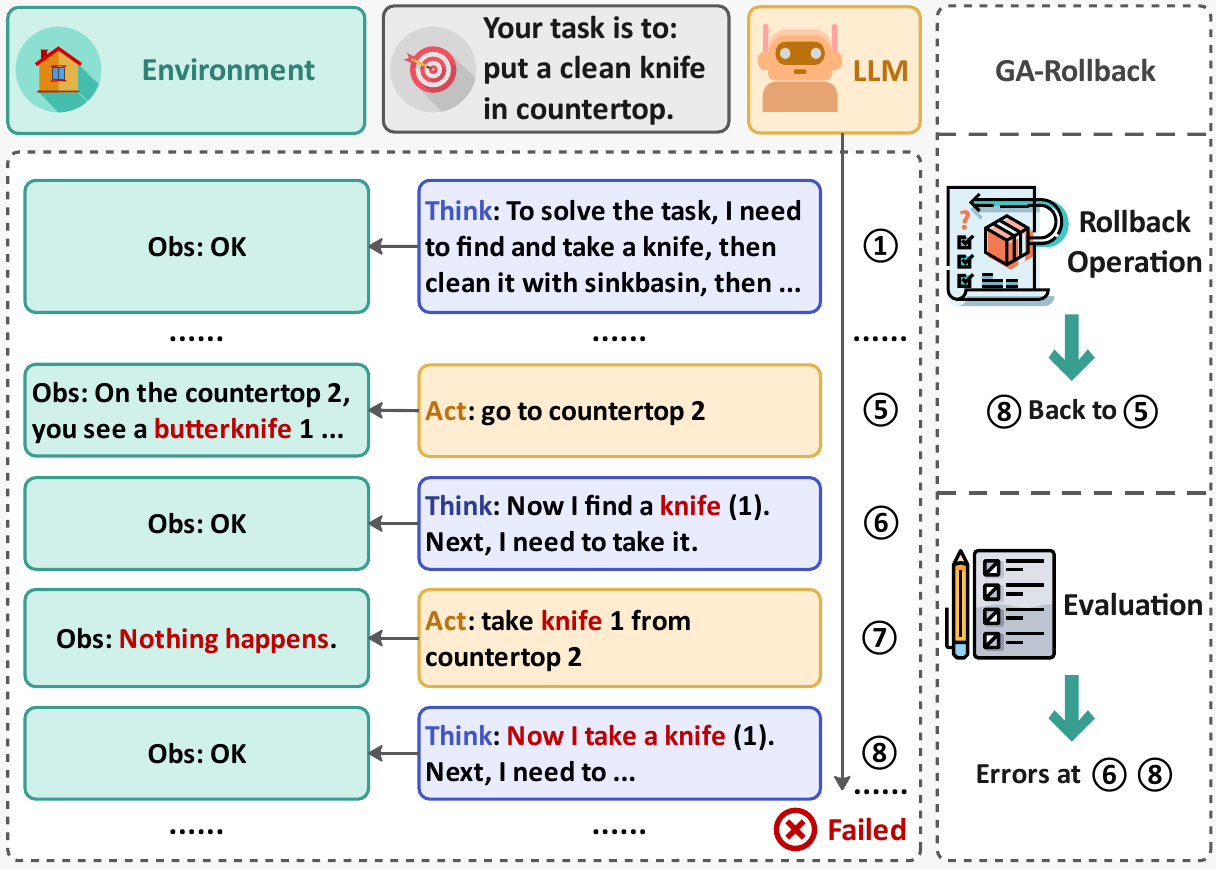}
    }
    \caption{An example of \textit{one-pass} paradigm. The trajectory is generated by LLaMA3.1-8B-Instruct in ReAct-style. Our approach blocks error propagation through rollback operations and ensures the quality of thinking processes through evaluation.}
    \label{fig:abs}
\end{figure}

Developing Large Language Model (LLM) agents capable of helping humans tackle real-world challenges has become a central focus in current artificial intelligence research \cite{survey1, survey2}.
Recently, researchers have leveraged the inherent self-planning abilities of LLMs to integrate thoughts with actions, enabling step-by-step reasoning processes during agent tasks \cite{song2022planner, react, qiao2024autoact}.
This enhancement empowers LLM agents to achieve remarkable improvements on diverse applications, including mathematical reasoning \cite{math, cobbe2021gsm8k}, web browsing \cite{webshop, mind2web, zhou2024webarena}, and embodied tasks \cite{alfworld, scienceworld}. 

However, this paradigm is inherently constrained by an \textit{one-pass} limitation, where each generated thought is inserted directly into the trajectory regardless of whether it is correct or not.
As a result, incorrect thoughts may persist in the context, influencing subsequent actions and ultimately compromising the outcome.
Take an example in Figure \ref{fig:abs} generated by LLaMA3.1-8B-Instruct in ReAct-style. 
In this scenario, the environment responds to all thoughts with a simple ``\textit{OK}'', which implies that the system assumes the correctness of these thoughts by default. 
Consequently, a low-quality thought that mistakes ``\textit{butterknife}'' for ``\textit{knife}'' leads to the invalid action ``\textit{take knife 1 from countertop 2}''.
To address this issue, many works have been devoted to improving the precision and clarity of reasoning processes.
Typically, self-correction methods \cite{self-refine, reflexion, critic} summarize refined plans using feedback from previous executions and improve the quality of the solution through multiple trials. 
Despite these achievements, the \textit{one-pass} reasoning pattern within each trial remains prone to error propagation due to its dependence on potentially flawed intermediate steps.

In this paper, we propose a novel framework called Generator-Assistant Stepwise Rollback (GA-Rollback) to induce better decision-making for goal-driven agents. 
Specifically, the generator interacts with the environment and supplies the assistant with essential contextual information.
Meanwhile, the assistant meticulously examines each action produced by the generator along with the corresponding observation.
When errors or suboptimal actions are detected, the assistant will provide detailed feedback to guide the generator in performing rollback operations, which revise previous incorrect actions to prevent error propagation.
To further enhance our framework, we introduce two key strategies: probability-based feedback evaluation (\S \ref{subsec:feedback}) to ensure feedback credibility, and Wait-Info strategy (\S \ref{subsec:wait-info}) to enrich contextual information in embodied tasks.
Experiments on three representative tasks reveal that our method outperforms several strong baselines across various models.
Moreover, our analysis demonstrates that GA-Rollback can function as a robust plug-and-play module, integrating seamlessly with other methods. In summary, our main contributions are as follows:

\begin{itemize}[leftmargin=*, topsep=1pt, itemsep=2.5pt, parsep=1pt]
\item We propose the GA-Rollback framework for llm-based agents, which separates action and thinking processes to ensure more precise and credible reasoning trajectory.

\item We introduce the probability-based feedback evaluation along with Wait-Info strategy designed for embodied environments.

\item Experiments on three tasks reveal that our method achieves notable improvement and exhibits stronger robustness, indicating its extensibility as a plug-and-play module.
\end{itemize}

\section{Related Work}
Our work is related to LLM agents and self-correction mechanism. 
In this section, we first review recent advances in LLM agents, followed by an analysis of self-correction mechanism.
\subsection{LLM Agents}
The reasoning and instruction-following capabilities that have emerged in LLMs \cite{LLM_survey1, LLM_survey2} make them capable enough to serve as intelligent agents to complete various tasks, such as web navigation \cite{mind2web, zhou2024webarena}, machine translation \cite{zhang2024paying, zuofei} and role playing \cite{tang2025rise}.
Standard LLM-agent methods employ structured reasoning frameworks to strengthen their analytical abilities \cite{tot, got}, or apply specialized decoding strategies to improve accuracy \cite{softsc}. Recent works have made significant breakthroughs by leveraging the inherent self-planning and reflection abilities of LLMs \cite{refiner, lats}. These methods achieve more effective interaction with external environments through task decomposition and planning within or between trials. 
While these methods have shown great progress, they still face a limitation in the lack of necessary evaluation of the reasoning process. This shortcoming can lead to the generation of low-quality reasoning steps, which may mislead subsequent actions and cause error propagation.

\subsection{Self-correction Mechanism}
Self-correction represents a feasible approach to improve responses from LLMs by enabling them to refine their outputs during inference \cite{self_correct1, self-refine}. 
The simplest implementation of self-correction prompts LLMs to provide feedback on their own responses and revise the responses based on the feedback \cite{prompt_selfcorrect}.
Recent studies have been dedicated to improving feedback by using additional resources, including external tools such as code executors \cite{self-debug, critic}, knowledge accessed through web browsing \cite{RAG, verify, check}, and observation gathered from simulation environments \cite{reflexion}.
Notably, \citet{selfcorrection_survey} identified feedback generation as the bottleneck in self-correction process, emphasizing that reliable feedback is essential for LLMs to successfully complete assigned tasks. 
Building upon these insights, our framework leverages LLMs as independent assistants in interactive environments and incorporates appropriate evaluations to enhance the credibility of generated feedback.

\begin{figure*}[pt]
    \centering
    \scalebox{0.976}{
    \includegraphics[width=\linewidth]{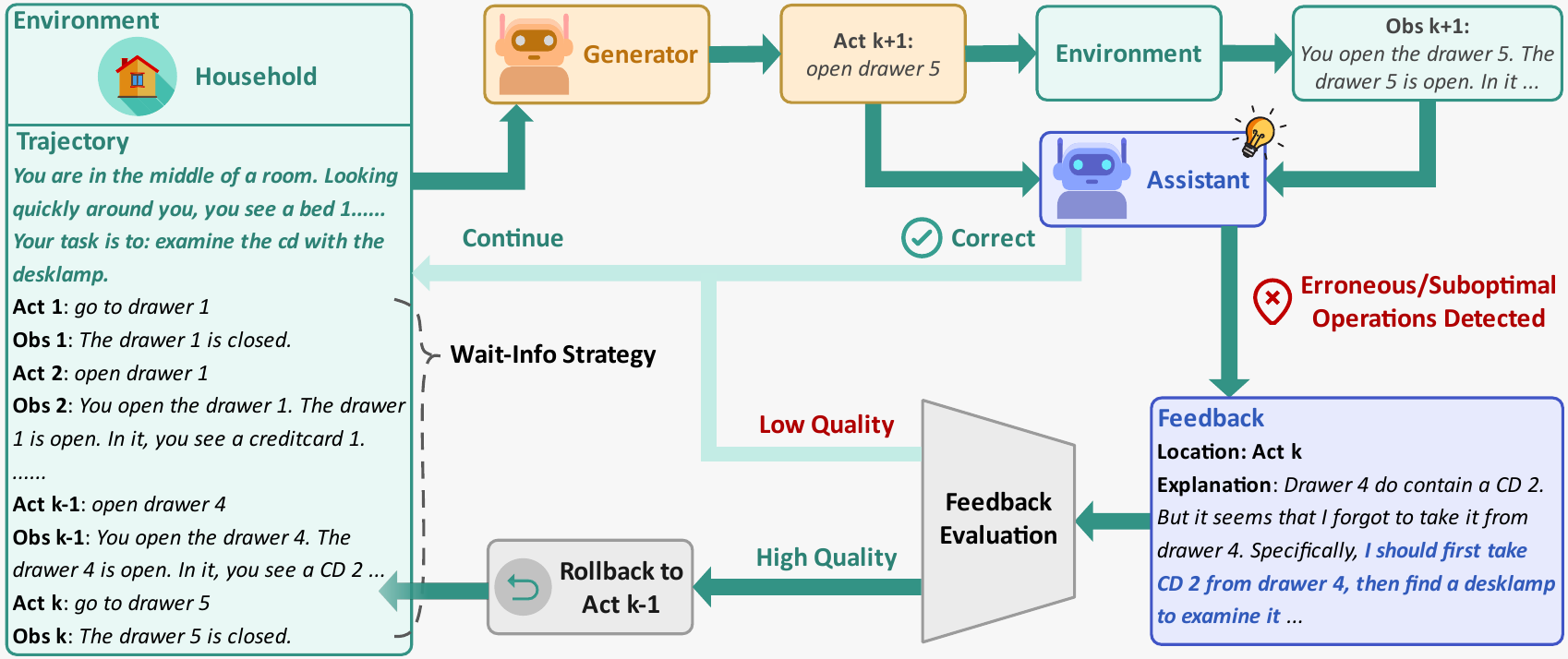}
    }
    \caption{An illustrative overview of GA-Rollback framework. The dark green arrows represent the current workflow. Each action and observation will be reviewed by the assistant, and the feedback provided by the assistant will also be evaluated before being applied to the trajectory.}
    \label{fig:method}
\end{figure*}

\section{Methodology}
The overall architecture of our method is illustrated in Figure \ref{fig:method}. In this framework, the generator produces actions along the trajectory, while the assistant serves as a rollback controller that provides detailed feedback upon the detection of errors or suboptimal actions (\S \ref{rollback}). To further enhance the performance of our framework, we propose two key strategies: probability-based feedback evaluation (\S \ref{subsec:feedback}) and Wait-Info strategy designed for embodied tasks (\S \ref{subsec:wait-info}).

We begin by formalizing the agent task and its trajectory. Given a pre-trained LLM denoted as $\mathcal{P}(\cdot)$, an agent needs to find a desired reasoning path toward addressing the given problem. We refer to the whole sequence of reasoning and the corresponding changes in the environment as a \textbf{trajectory} $\mathcal{T}_n = [\mathcal{I}, o_0, a_1, o_1, ..., a_n, o_n]$, where $\mathcal{I}$ is the task description, $n$ is the length of the trajectory, $o_0$ denotes the initial observation of the environment, $a_{1...n}$ are actions made by the agent. Under normal circumstances, the intermediate action $a_n$ is generated as $a_n \sim \mathcal{P}(a_n| \mathcal{T}_{n-1})$, and the current action set is defined as $\mathcal{A}_n = [a_1, a_2, ..., a_n]$. When an erroneous action, denoted as $\widehat{a}_t$, is retained within the trajectory $\mathcal{T}_t$, it is likely to lead to error propagation. Intuitively, the task cannot be accomplished through the erroneous action set $\widehat{\mathcal{A}}_n = [a_1, a_2, ..., \widehat{a}_t, \widehat{a}_{t+1}, ..., \widehat{a}_n]$.

To alleviate the impact of erroneous actions, self-correction mechanism is usually incorporated into the standard framework, whereby an optimization process is implemented after each step or each trial. In the scenario of stepwise self-correction, the agent adjusts its behavior based on current trajectory $\mathcal{T}_n$ and the feedback $f_n$ from optimizer: $a_{n+1} \sim \mathcal{P}(a_{n+1}| f_n, \mathcal{T}_n)$. However, the delayed detection of errors leaves incorrect actions in the trajectory, which may adversely affect subsequent steps. Therefore, we introduce the concept of \textbf{rollback} \cite{roll} into self-correction. When error occurs, LLM agents are required to create better reasoning path by continuously rolling back from current trajectory $\mathcal{T}_n$ to prior $\mathcal{T}_m$, where $m \in [0, n-1]$. 

\subsection{Action Rollback based on Feedback}
\label{rollback}
In our framework, we employ an LLM as the assistant to support the generator engaged in the task.
Concretely, each time the action $a_n$ generated by the generator and the corresponding observation $o_n$ are updated in the trajectory $\mathcal{T}_n$, the assistant analyzes $\mathcal{T}_n$ in a chain-of-thought \cite{cot} manner. For tasks involving typically long trajectories, we enable the assistant review the steps from back to front to efficiently locate the most recent error encountered.
If the assistant detects errors or suboptimal actions in $\mathcal{T}_n$ during analysis, it will provide feedback $f_n$ including identified erroneous actions along with specific explanations. 
We assume that the identified erroneous actions are $[\widehat{a}_t, \widehat{a}_{t+1}, ..., \widehat{a}_n]$ with $\widehat{a}_t$ being the earliest error, and the number of rollback steps is $N = n-t+1$. 
Considering that an excessively large rollback span may result in the omission of certain details, we define an upper bound $\overline{N}$ for $N$. When $N$ exceeds $\overline{N}$, it will be replaced by $\overline{N}$ to ensure a more precise and meticulous rollback operation.

Upon receiving the assistant's instruction to rollback to $\mathcal{T}_{t-1}$, we first revert the state of the environment. Specially, we retrieve the action set $\mathcal{A}_{t-1} = [a_1, ..., a_{t-1}]$ from $\mathcal{T}_{t-1}$, reset the environment, and sequentially execute the actions in  $\mathcal{A}_{t-1}$ to restore the state to that of $\mathcal{T}_{t-1}$.
Once this process is completed, the generator produces a new action $a_t$ based on $\mathcal{T}_{t-1}$ and $f_n$.
By rolling back to the previous state, the generator can continue its search iteratively until it finds a feasible path to complete the task. 
Drawing inspiration from \citet{roll}, we also incorporate past mistakes as experiential knowledge into the prompt to avoid the repetition of similar errors.

\subsection{Probability-based Feedback Evaluation}
\label{subsec:feedback}
Since feedback serves as the trigger for the rollback operation, its quality is of vital importance to the overall performance.
If low-quality feedback intervenes in the trajectory, issues such as unnecessary rollbacks may arise. 
Based on the findings of \citet{softsc}, we employ a probability-based evaluation to measure the confidence of the feedback generated by the assistant, thereby mitigating this adverse effect.

For a given output $Y = \{y_1, ..., y_m\}$ generated by the assistant, we score its confidence using the mean-pooled probability of its tokens:
\begin{equation}
    \mathrm{Score_Y} = \frac{1}{m}\sum_{i=1}^{m}\mathcal{P}(y_i|y_{<i}, X),
\end{equation}
where $X$ denotes the input containing task description and current trajectory. If $\mathrm{Score_Y}$ is lower than a predefined threshold $\theta$, the corresponding feedback will be discarded. In such cases, the feedback will neither be passed to the generator nor trigger a rollback operation. To determine the threshold $\theta$, we first analyze the distribution pattern of this confidence metric, then empirically select an appropriate value through systematic testing at certain intervals across the distribution range (\S \ref{subsec:thre}).

\subsection{Wait-Info Strategy for Embodied Tasks}
\label{subsec:wait-info}
In embodied environments, agents need to meticulously check multiple locations and interact with various objects before discovering a viable solution to accomplish the assigned task.
These exploratory behaviors represent necessary steps in the task-solving process.
Thus, premature intervention during this exploration phase, particularly from the assistant, may be counterproductive and lead to redundant rollback operations, as the assistant might over-analysis these exploratory behaviors.

To resolve the aforementioned problem, our Wait-Info strategy restricts the assistant's involvement in the early stages, allowing the generator to independently generate and execute actions without immediate intervention. 
The trajectory analysis by the assistant is permitted only after the generator completes the generation and execution of $k$ consecutive actions in the environment.
If the rollback operation reduces the trajectory length below $k$, the assistant will remain locked until the generator extends the trajectory length back to $k$. 
This strategy enables the generator to explore its surroundings more thoroughly, providing the assistant with a richer set of information $\mathcal{T}_{\geq k}$ for decision-making.

\section{Experiments}
\subsection{Experimental Settings}
\noindent \textbf{Tasks.} We evaluate our framework on three representative agent tasks: Game of 24 \cite{tot} for mathematical reasoning, ALFWorld \cite{alfworld} for embodied house hold tasks, Webshop \cite{webshop} for web navigation. 
Webshop provides a dense reward ranging from 0 to 1 to quantify task completion, while Game of 24 and ALFWorld only provide binary rewards to indicate whether the task is completed. 
We made slight adjustments to the Game of 24 task to ensure that the agent receives relevant observation for each executed action. During evaluation, we sample 500 instances each from Webshop and Game of 24, and use the out-of-distribution test set of ALFWorld.
More details can be found in Appendix \ref{app:task_details}.

\noindent \textbf{Baselines.} We compare the GA-Rollback framework with several agent methods that are known for their ability to generalize across diverse tasks. 
(1) \textbf{Few-shot} \cite{few-shot} (referred to as \textbf{Act-only} in the table) provides the model with a set of examples to capture task-specific patterns. 
(2) \textbf{CoT} \cite{cot} guides the model to explicitly generate intermediate rationales before reaching the final answer.
(3) \textbf{ReAct} \cite{react} enhances task completion by interleaving reasoning with actions during execution, where the model autonomously generates thinking processes based on the observation to refine subsequent actions. 
(4) \textbf{Reflexion} \cite{reflexion} synthesizes more precise plans between attempts to optimize the trajectory of subsequent trials. 
(5) \textbf{ReAct + Reflexion} combines both methods for more comprehensive thinking and planning. 
Furthermore, we integrate GA-Rollback with these methods to investigate its extensibility.

\begin{table*}[ht]
    \centering
    \scalebox{0.816}{
    \begin{tabular}{cl|w{c}{4.5em}w{c}{4.5em}w{c}{4.5em}w{c}{4.5em}|w{c}{4.5em}}
    \toprule
    \multirow{2}{*}{\textbf{Model}} & \multicolumn{1}{c|}{\multirow{2}{*}{\textbf{Method}}} 
    & \textbf{Game of 24} 
    & \textbf{ALFWorld} 
    & \multicolumn{2}{c|}{\textbf{Webshop}} 
    & \multirow{2}{*}{\textbf{Avg. SR}} \\
    & & SR & SR & Reward & SR \\
    \midrule
    \multirow{8}{*}{\textit{LLaMA-3.1-8B-Instruct}} & Act-only & 0.4 & 33.6 & 62.3 & 30.6 & 21.5 \\
    & CoT & 0.0 & 32.8 & 47.8 & 27.4 & 20.1 \\
    & ReAct & 0.4 & 22.4 & 47.9 & 27.0 & 16.6 \\
    & Reflexion & 5.2 & \textbf{44.8} & \underline{63.3} & 32.8 & \underline{27.6} \\
    & ReAct + Reflexion & \underline{6.4} & 26.1 & 54.2 & 30.2 & 20.9 \\
    \cmidrule(lr){2-7}
    & \textbf{GA-Rollback} & 4.2 & 38.8 & 61.1 & \underline{34.0} & 25.7 \\
    & \textbf{GA-Rollback} + ReAct & \underline{6.4} & 18.7 & 44.8 & 25.6 & 16.9 \\
    & \textbf{GA-Rollback} + Reflexion & \textbf{6.8} & \underline{42.5} & \textbf{68.1} & \textbf{40.6} & \textbf{30.0} \\
    \midrule
    \midrule
    \multirow{8}{*}{\textit{GLM4-9B-Chat}} & Act-only & 0.0 & 73.1 & 65.3 & 32.6 & 35.2 \\
    & CoT & 0.0 & 70.9 & 64.8 & 34.8 & 35.2 \\
    & ReAct & 0.2 & 76.8 & 56.7 & 23.0 & 33.3 \\
    & Reflexion & \underline{7.2} & 77.6 & \underline{67.3} & 33.6 & 39.5 \\
    & ReAct + Reflexion & 6.4 & \textbf{91.8} & 59.9 & 26.0 & \underline{41.4} \\
    \cmidrule(lr){2-7}
    & \textbf{GA-Rollback} & 5.4 & 78.4 & 66.9 & \underline{37.8} & 40.5 \\
    & \textbf{GA-Rollback} + ReAct & 4.6 & 80.6 & 37.0 & 15.6 & 33.6 \\
    & \textbf{GA-Rollback} + Reflexion & \textbf{9.6} & \underline{85.8} & \textbf{70.2} & \textbf{41.4} & \textbf{45.6} \\
    \midrule
    \midrule
    \multirow{8}{*}{\textit{Qwen2.5-14B-Instruct}} & Act-only & 5.4 & 78.3 & 60.7 & 34.0 & 39.2 \\
    & CoT & 7.0 & 86.6 & 57.8 & 36.2 & 43.3 \\
    & ReAct & 7.6 & 89.6 & 43.6 & 26.6 & 41.3 \\
    & Reflexion & 18.8 & 91.8 & \underline{68.5} & 39.6 & \underline{50.1} \\
    & ReAct + Reflexion & 18.8 & \textbf{95.5} & 47.8 & 29.4 & 47.9 \\
    \cmidrule(lr){2-7}
    & \textbf{GA-Rollback} & 17.2 & 89.6 & 65.7 & \underline{41.2} & 48.8 \\
    & \textbf{GA-Rollback} + ReAct & \underline{19.4} & 83.6 & 36.9 & 23.8 & 42.3 \\
    & \textbf{GA-Rollback} + Reflexion & \textbf{23.6} & \underline{92.6} & \textbf{70.4} & \textbf{45.2} & \textbf{53.8} \\
    \bottomrule
    \end{tabular}
    }
    \caption{Performance of different methods on three agent tasks. The best results of each model are marked in \textbf{bold} and the second-best results are marked with \underline{underline}. SR: success rate.}
    \label{tab:main}
\end{table*}
\noindent \textbf{Implementation Details.} We conduct our experiments on LLaMA-3.1-8B-Instruct \cite{llama3.1}, GLM4-9B-Chat \cite{glm2024chatglm}, and Qwen2.5 series \cite{qwen2.5} to provide comprehensive results. 
Additionally, results for models of varying scales, such as Qwen2.5-3B-Instruct, are provided in Appendix \ref{app:more_model_results}.
All experiments are carried out on NVIDIA HGX H20 96G GPUs. 
In the main experiments, we use the same model as generator and assistant. 
We set the ``max\_new\_tokens'' for the generator to 100 and for the assistant to 500, with both having a ``temperature'' of 0.1. 
We also leverage dynamic model compilation to accelerate inference.
The number of Wait-Info steps is set to 6, and the quality threshold $\theta$ for feedback evaluation is set to 0.93.
The number of trials for Reflexion is set to 2. 
To prevent unlimited rollback operations from causing an infinite loop, we set the maximum number of rollback attempts per task at 6.

\subsection{Main Results}
Table \ref{tab:main} presents the performance comparison of different methods on three agent tasks. Based on the results, we have the following findings: 

(i) When applied independently, GA-Rollback outperforms several baselines in complex reasoning tasks. For instance, in the Game of 24 benchmark, GA-Rollback alone achieves success rate of 4.2\% (LLaMA-3.1-8B-Instruct), 5.4\% (GLM4-9B-Chat), and 17.2\% (Qwen2.5-14B-Instruct), representing 10.5×, 27.0×, and 2.3× improvements over ReAct. This indicates the effectiveness of our GA-Rollback as a dynamic component in multi-step reasoning.

(ii) The synergistic potential of GA-Rollback is particularly evident when combined with Reflexion. From the results of Webshop, the success rate of GA-Rollback is higher than that of Reflexion, but its reward is lower. This suggests that in the absence of an overall plan, GA-Rollback may cause the agent to miss some necessary intermediate steps. Combining these two methods could leverage their complementary strengths, leading to enhanced performance in agent tasks. 

(iii) Even when GA-Rollback is integrated with Reflexion, it attains suboptimal performance on ALFWorld. Such outcomes indicate that embodied tasks involving longer trajectories remain challenging for GA-Rollback, particularly in determining which step to rollback to. 

\begin{figure*}[pt]
    \centering
    \begin{subfigure}[b]{0.465\textwidth}
        \includegraphics[width=\textwidth]{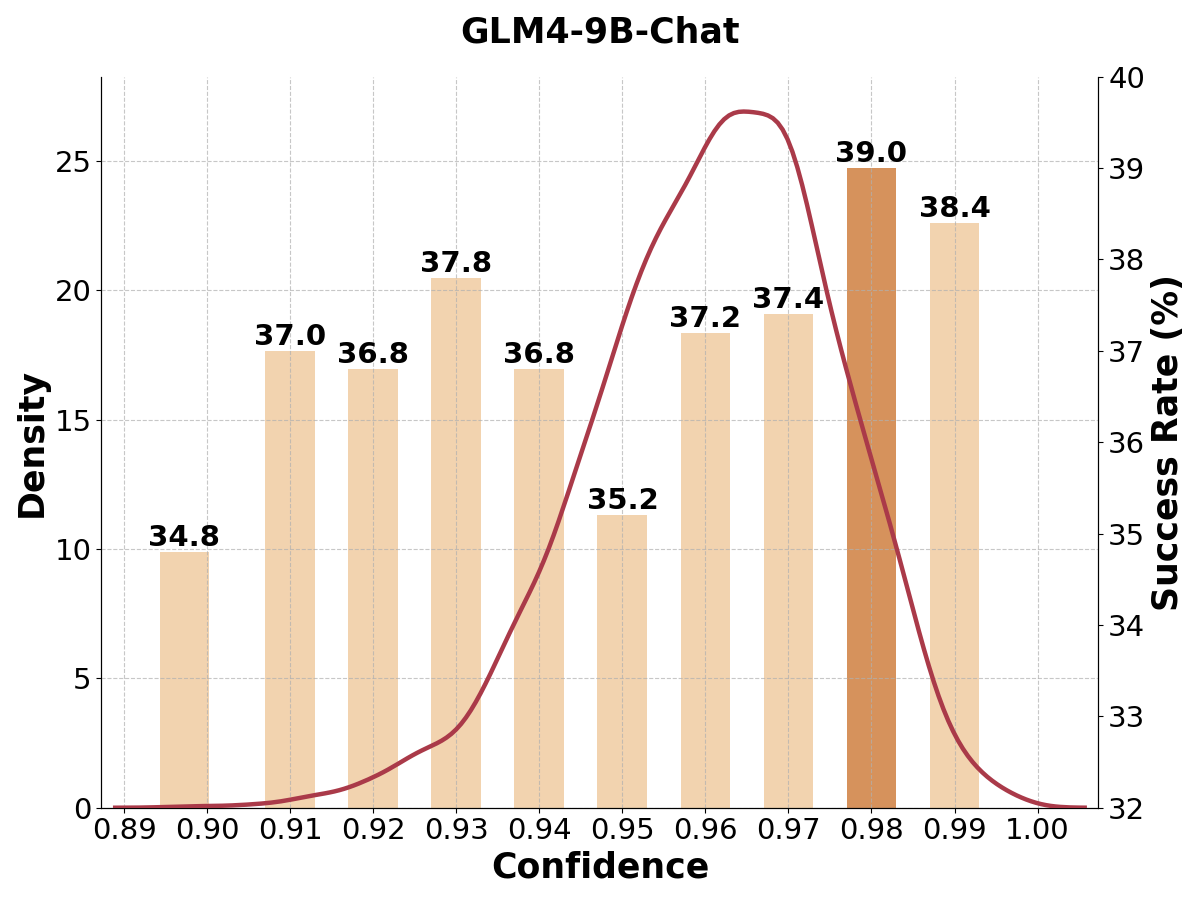}
        \caption{Results of GLM4-9B-Chat on Webshop}
        \label{fig:subfig1}
    \end{subfigure}
    \hspace{0.03\textwidth} 
    \begin{subfigure}[b]{0.465\textwidth}
        \includegraphics[width=\textwidth]{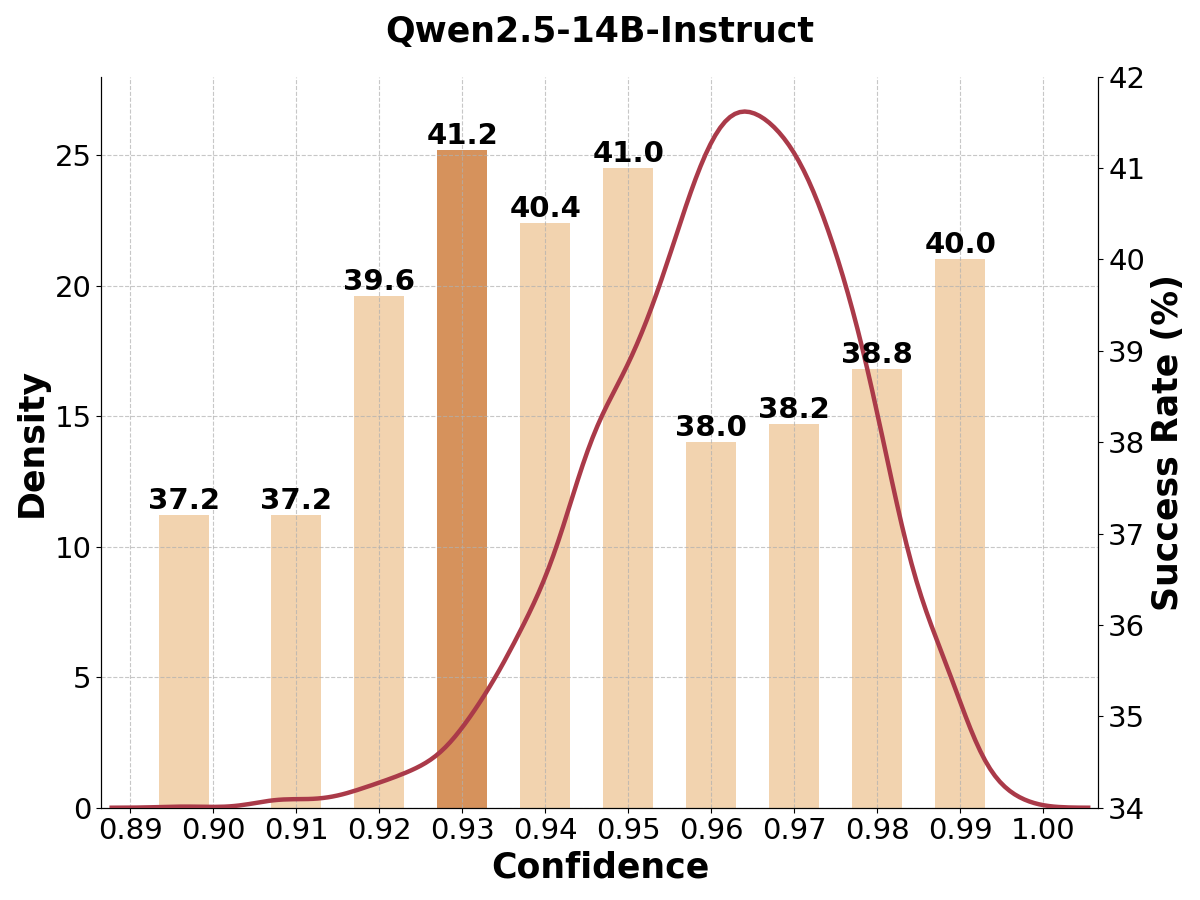}
        \caption{Results of Qwen2.5-14B-Instruct on Webshop}
        \label{fig:subfig2}
    \end{subfigure}
    
    \vspace{0.005\textwidth}
    \begin{subfigure}[b]{0.465\textwidth}
        \includegraphics[width=\textwidth]{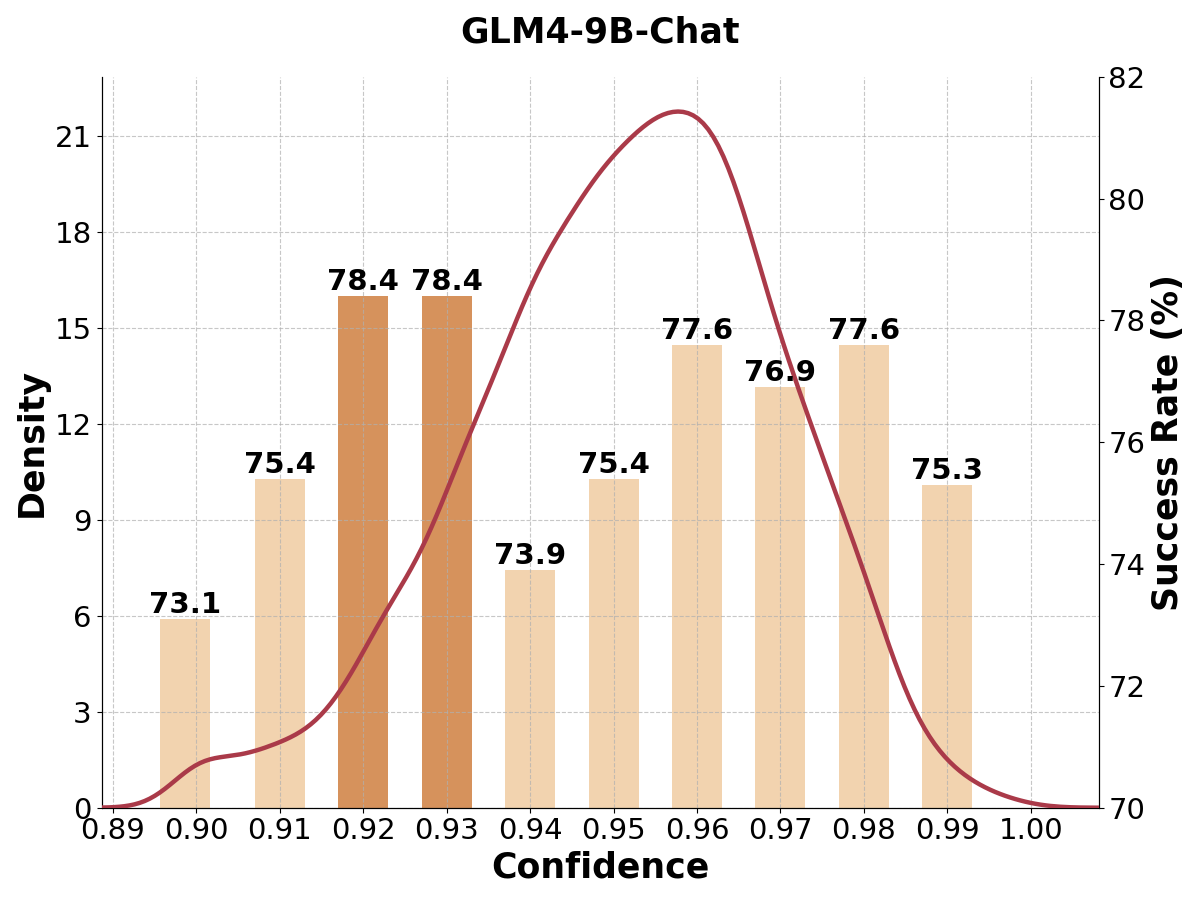}
        \caption{Results of GLM4-9B-Chat on ALFWorld}
        \label{fig:subfig3}
    \end{subfigure}
    \hspace{0.03\textwidth} 
    \begin{subfigure}[b]{0.465\textwidth}
        \includegraphics[width=\textwidth]{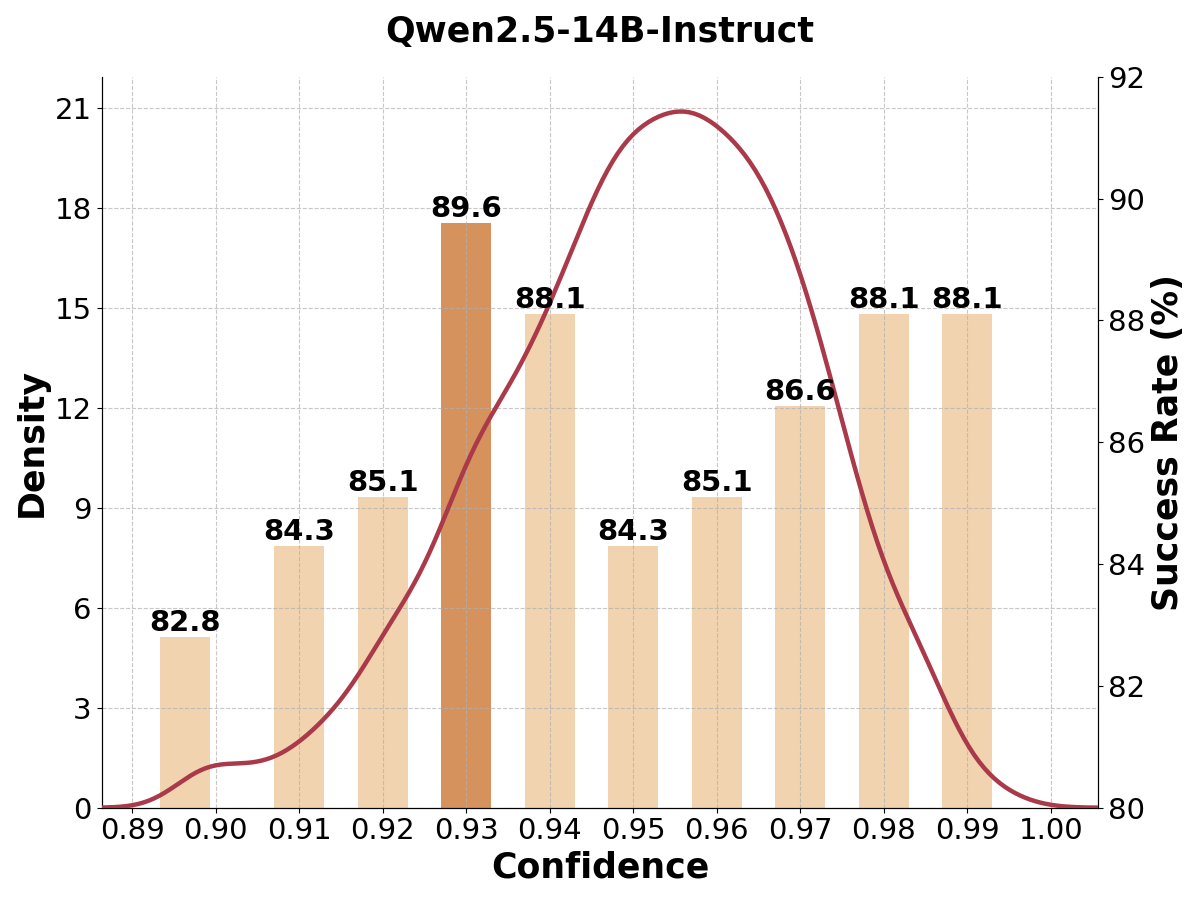}
        \caption{Results of Qwen2.5-14B-Instruct on ALFWorld}
        \label{fig:subfig4}
    \end{subfigure}
    \caption{The kernel density estimation curve (red) and the success rate (orange) as a function of threshold $\theta$. The leftmost bar in each subgraph corresponds to the performance without feedback evaluation. The darker colored bars highlight the respective performance peaks.}
    \label{fig:thre}
\end{figure*}

(iv) Surprisingly, we observe a performance decline in all three models when conducting Webshop in ReAct-style. Besides, while ReAct improves the ALFWorld success rate for Qwen2.5-14B-Instruct, it results in a 33\% performance drop for LLaMA3.1-8B-Instruct. This demonstrates that the effectiveness of ReAct is contingent on the reasoning capabilities of LLMs. Incorrect reasoning caused by ReAct can adversely affect the outcomes. Moreover, we observe that the thinking processes generated by GA-Rollback and ReAct may conflict with each other, leading to instability when combining the two methods. Appendix \ref{app:cases} presents an example of such conflict.

\subsection{Analysis of Feedback Evaluation}
\label{subsec:thre}

As mentioned in section \ref{subsec:feedback}, we utilize mean-pooling operations on the assistant's output tokens to represent its overall confidence in the generated feedback. In this section, we investigate the distribution of this confidence metric and its impact on the GA-Rollback framework. 

Firstly, we implement the GA-Rollback framework using GLM4-9B-Chat and Qwen2.5-14B-Instruct to conduct Webshop and ALFWorld tasks, during which we collect the feedback confidence scores. As shown in Figure \ref{fig:thre}, the kernel density estimation curve of confidence approximately follows a normal distribution pattern. This indicates that the model maintains relatively stable confidence levels across different task instances without exhibiting extreme overconfidence or underconfidence tendencies. 

Based on the observed distribution, we leverage a threshold $\theta$ to filter out feedback with confidence falling below this value. To determine an optimal filtering criterion, we vary the threshold at 0.01 intervals across the distribution range and evaluate the corresponding success rates. As illustrated in Figure \ref{fig:thre}, without implementing feedback evaluation, the success rates of the two models on both tasks are relatively low. We also observe two distinct optimal threshold regions for feedback filtering: A threshold of 0.93 effectively eliminates low-confidence noise while preserving the majority of feedback; a stricter threshold of 0.98 retains only the highest-confidence feedback. 
We ultimately adopt 0.93 for two considerations:

\begin{itemize}[leftmargin=*, topsep=1pt, itemsep=2.5pt, parsep=1pt]
\item \textbf{Generalizability}: The threshold of 0.93 yields near-optimal results for both models.

\item \textbf{Efficiency}: The threshold of 0.98 results in a significant portion of generated feedback being discarded, leading to unnecessary computational overhead and reduced efficiency.
\end{itemize}

\subsection{Analysis of Wait-Info Strategy}

\begin{figure}[pht]
    \centering
    {
    \includegraphics[width=\linewidth]{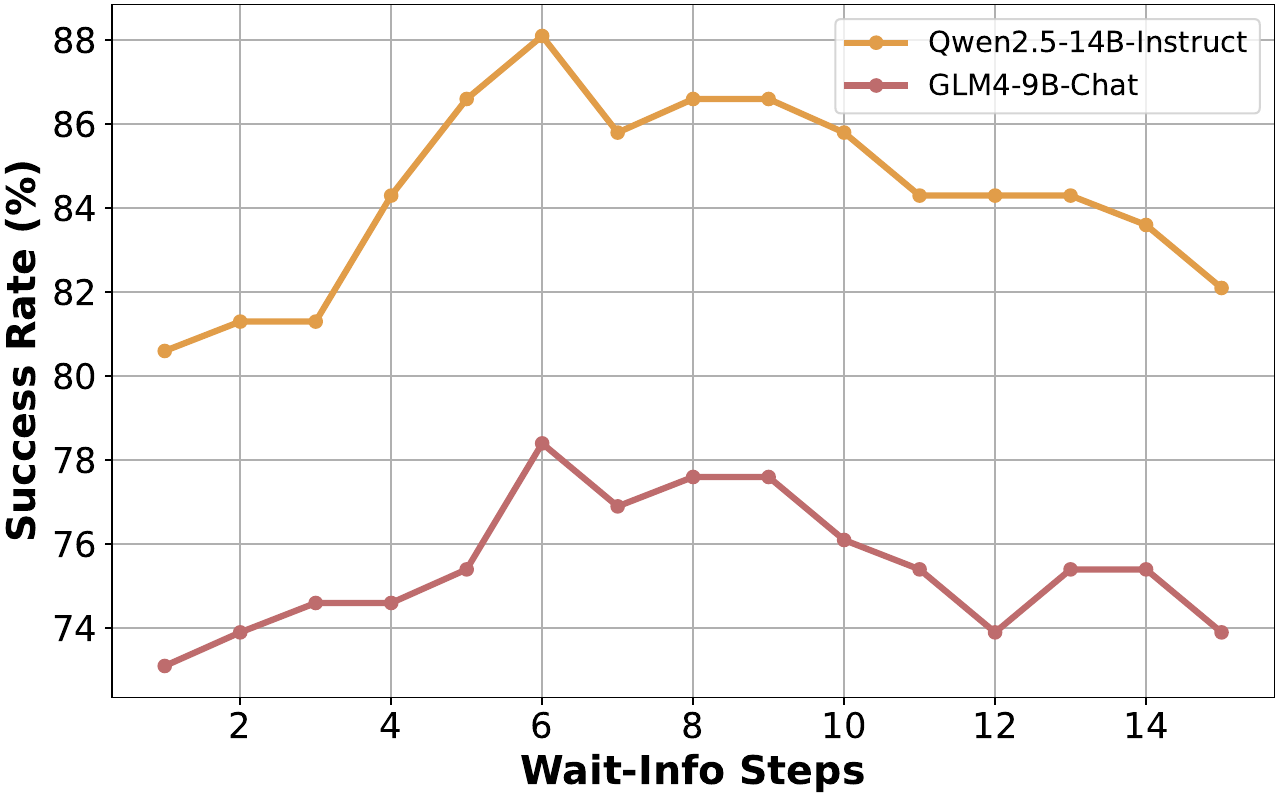}
    }
    \caption{The performance of GA-Rollback on ALFWorld with different Wait-Info steps.}
    \label{fig:wait-info}
\end{figure}

Figure \ref{fig:wait-info} illustrates how the \textbf{Wait-Info} strategy—the minimum number of generator actions delivered to the assistant—affects GLM-4-9B-Chat and Qwen-14B-Instruct on ALFWorld. Performance follows a clear $\cap$-shaped curve: success rate climbs steadily until $k$$\approx$6 (yielding a 4-8 percentage-point gain over the no-wait baseline) and then drops. A moderate horizon enriches the assistant’s context with a temporally coherent exploration burst, making causal chains explicit and exposing discrepancies between intended and executed actions; too small a horizon starves the assistant of signal, while too large a horizon floods it with noisy, combinatorial states.

This pattern reveals a principled \textbf{information-bandwidth trade-off} rather than a mere implementation tweak. Wait-Info functions as a horizon-gated scheduler that clearly separates the generator’s high-entropy exploration phase from the assistant’s low-entropy diagnosis phase. Up to the optimal $k^*$, progressive context enrichment helps the assistant perform targeted rollbacks; beyond $k^*$, the cognitive burden of analyzing long, error-diffused traces outweigh the benefit of deeper look-ahead, producing diminishing returns. Viewed in this light, Wait-Info becomes a reusable design knob for cooperative LLM paradigms, and we expect the same $\cap$-shaped relationship to generalize to other agent benchmarks where partial thinking trajectories must be exchanged.

\subsection{Comparison with Tree-search Method}

We further compare our approach with tree-search method ToT-BFS \cite{tot} on Game of 24. As shown in Table \ref{tab:comp_tree}, GA-Rollback achieves greater performance improvements while resulting in only 2× token cost and 1.5× runtime overhead compared to ToT-BFS.

\begin{table}[ph]
    \centering
    \scalebox{0.63}{
    \begin{tabular}{p{2.2cm}l|ccc}
        \toprule
        \textbf{Model} & \textbf{Method} & \textbf{SR} & \textbf{Avg. Token (K)} & \textbf{Avg. Time (s)} \\
        \midrule
        \multirow{3}{=}{\textit{LLaMA3.1-8B-Instruct}} & Act-only & 1.0 & 10.7 & 21.1 \\
        & ToT-BFS     & 1.2 & 22.8 & 85.2 \\
        & \textbf{GA-Rollback} & 3.2 & 38.8 & 112.3 \\
        \midrule
        \multirow{3}{=}{\textit{GLM4-9B-Chat}} & Act-only & 0.0 & 10.9 & 27.2 \\
        & ToT-BFS     & 1.6 & 18.0 & 83.8 \\
        & \textbf{GA-Rollback} & 6.0 & 36.4 & 123.0 \\
        \midrule
        \multirow{3}{=}{\textit{Qwen2.5-14B-Instruct}} & Act-only & 5.8 & 9.8 & 29.3 \\
        & ToT-BFS     & 6.4  & 17.7 & 82.2 \\
        & \textbf{GA-Rollback} & 18.4 & 35.4 & 148.2 \\
        \bottomrule
    \end{tabular}}
    \caption{Performance comparison with tree-search method ToT-BFS \cite{tot} on Game of 24.}
    \label{tab:comp_tree}
\end{table}

Notably, our approach addresses two critical limitations of tree-search methods: (1) Although methods like ToT optimize stepwise decisions through scoring mechanisms, they remain inherently susceptible to error propagation. In contrast, our approach actively mitigates error propagation through rollback-driven self-correction, ensuring more robust performance. (2) In complex environments (e.g., embodied settings), the action space grows exponentially, leading to extremely high search complexity. Instead of exhaustive search, our approach emphasizes dynamic plan adjustment via rollback during exploration.

\begin{figure*}[pt]
    \centering
    \scalebox{0.978}{
    {
    \includegraphics[width=\linewidth]{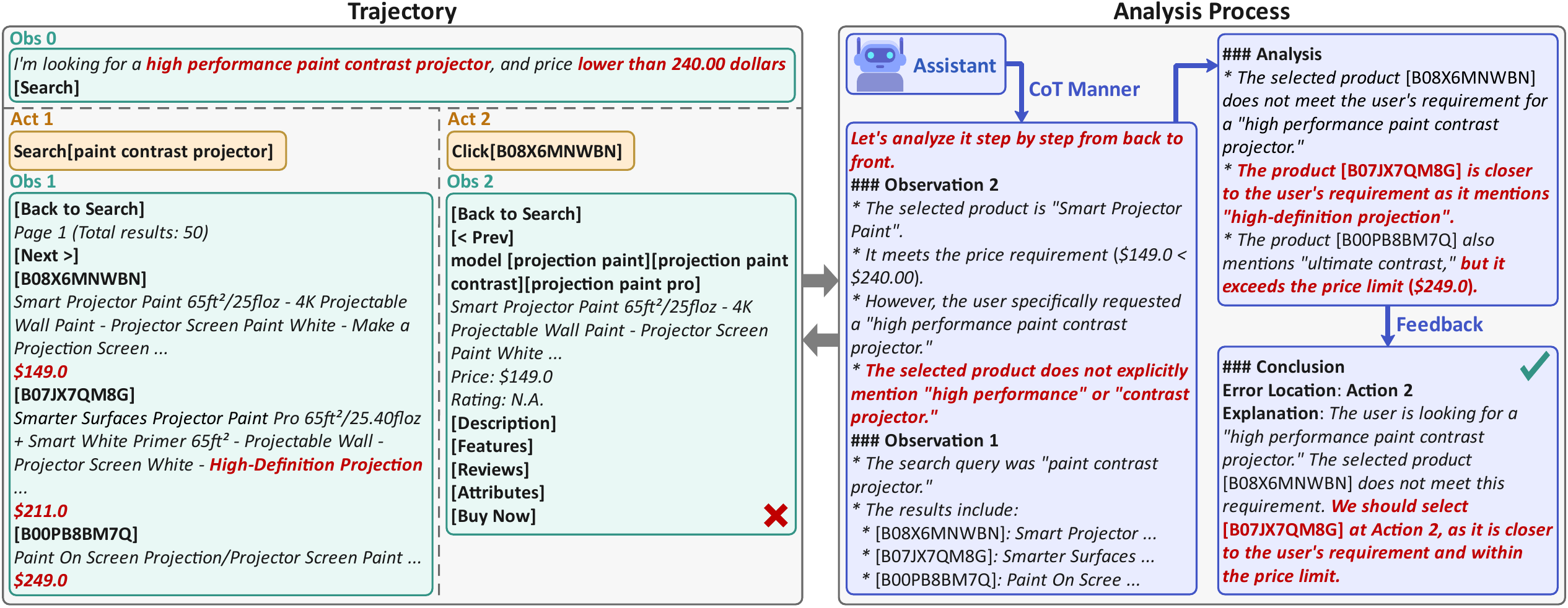}
    }
    }
    \caption{An example of trajectory analysis conducted by the assistant, with key information highlighted in red.}
    \label{fig:case}
\end{figure*}

\subsection{Ablation Study}
We remove feedback evaluation, Wait-Info strategy, and even the entire assistant from our framework to validate their individual contributions. 
As shown in Table \ref{tab:assist_ablation}, the removal of each component leads to a clear drop in overall success rate, underscoring the effectiveness of our method. 
Specifically, eliminating the assistant reduces the entire framework to an Act-only form, resulting in a significant performance decline, which highlights the crucial role of the assistant.

\begin{table}[pt]
\centering
\scalebox{0.82}{
\begin{tabular}{lccc}
\toprule
\textbf{Setting} & \textbf{Game of 24} & \textbf{ALFWorld} & \textbf{Webshop} \\
\midrule
\rowcolor{mygray}
\multicolumn{4}{l}{\textit{\textbf{14B as Generator and Assistant}}} \\
\textbf{GA-Rollback }      & \textbf{17.2} & \textbf{88.1} & \textbf{41.2} \\
\textbf{w/o} evaluation    & 15.4 & 82.8 & 37.2 \\
\textbf{w/o} Wait-Info    &  -   & 80.6 & - \\
\textbf{w/o} Assistant     & 5.4  & 78.3 & 34.0 \\
\midrule
\rowcolor{mygray}
\multicolumn{4}{l}{\textit{\textbf{Different Assistants}}} \\
0.5B             & 6.8 & 82.8 & 38.4 \\
3B               & 9.4 & 85.8 & 39.2 \\
7B               & 10.2 & 86.6 & 39.6 \\
72B              & \textbf{19.4} & \textbf{90.3} & \textbf{44.0} \\

\bottomrule
\end{tabular}
}
\caption{The ablation study on each component of GA-Rollback and assistants. The generator is consistently set to Qwen2.5-14B-Instruct, and the assistant is selected from Qwen2.5 Instruct models.}
\label{tab:assist_ablation}
\end{table}

To further study the impact of the assistant, we keep the generator fixed and experiment with Qwen2.5 Instruct models of varying scales (ranging from 0.5B to 72B) as the assistant.
Notably, as the scale of the assistant increases, its reasoning capability improves correspondingly. This enhancement allows the assistant to analyze the trajectories more thoroughly and provide refined feedback, thereby boosting success rate.

\subsection{Case Study}
Here, we present an example of the trajectory analysis conducted by the assistant on Webshop. As shown in Figure \ref{fig:case}, the user needs a ``\textit{high performance paint contrast projector}'' below \$240. Initially, the generator selects a product that appears to meet these requirements. However, after a thorough examination of the trajectory, the assistant discerns that the chosen item does not satisfy the user's requirement for ``\textit{high performance}''. Recognizing this issue, the assistant revisits the search results and excludes the product ``[B00PB8BM7Q]'', which matches the description but exceeds the budget.
Then, it selects the appropriate alternative ``[B07JX7QM8G]'' and provides detailed feedback to inform the generator.
This meticulous verification and subsequent rollback operation enable the generator to more effectively achieve its task objectives. More detailed examples are provided in Appendix \ref{app:cases}.

\subsection{Multi-Trial Analysis of GA-Rollback + Reflexion}

\begin{figure*}[pt]
    \centering
    \scalebox{0.945}{
    {
    \includegraphics[width=\linewidth]{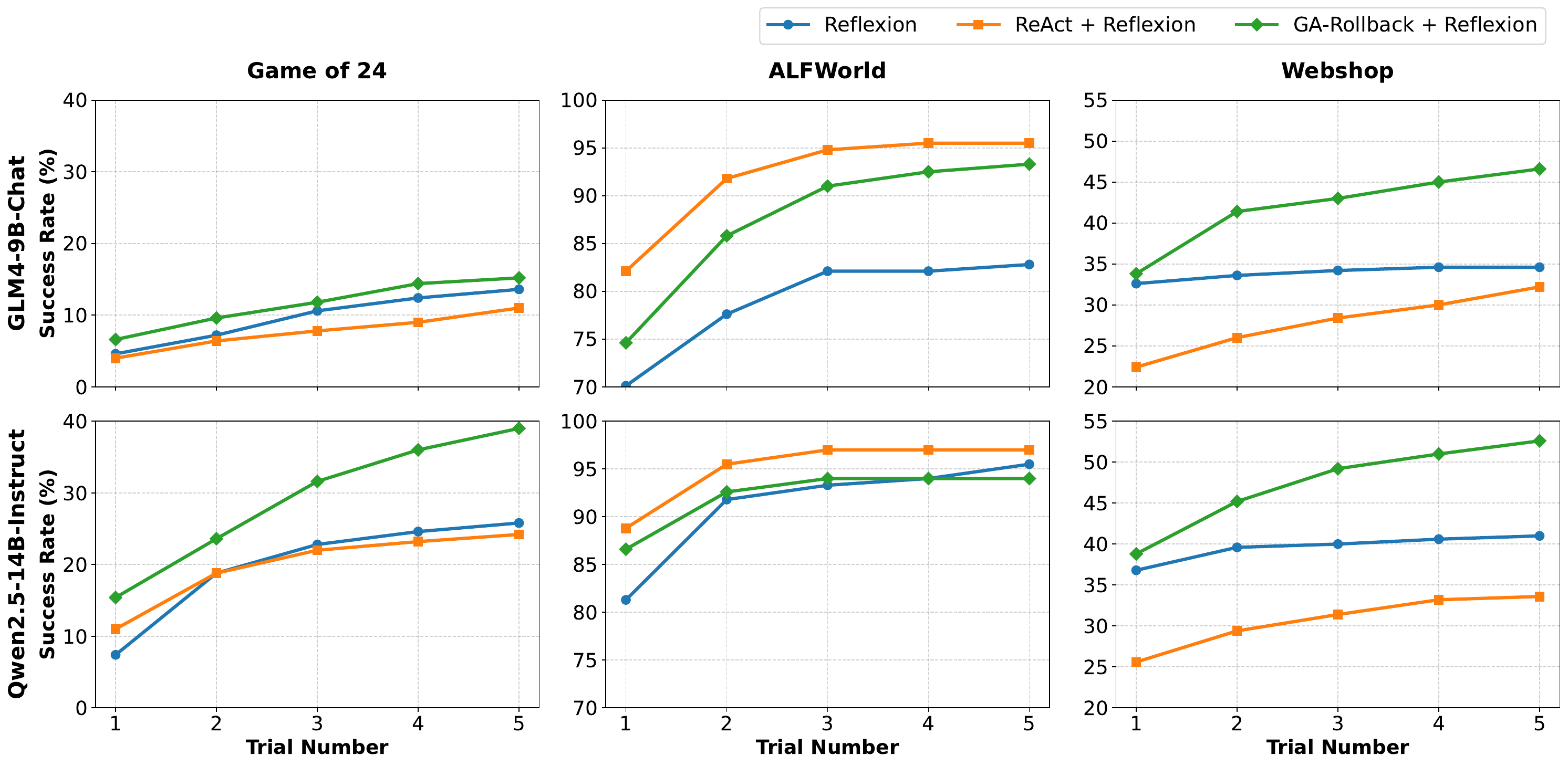}
    }
    }
    \caption{Comparison of success rate across trials for Reflexion, ReAct+Reflexion, and GA-Rollback+Reflexion.}
    \label{fig:rollback+reflex}
\end{figure*}

To further investigate the compatibility of our GA-Rollback and Reflexion, we evaluate its success rate across multiple rounds of trials on three agent tasks. The results, as illustrated in Figure \ref{fig:rollback+reflex}, reveal that the success rate of GA-Rollback + Reflexion exhibits a steady increase across three tasks with the rise in trial numbers. Notably, this approach significantly outperforms other multi-trial paradigms on both Game of 24 and Webshop, underscoring the versatility of GA-Rollback as a plug-and-play component that can be seamlessly integrated with other methods. In contrast to the findings reported by \citet{reflexion}, smaller-scale LLMs such as GLM4-9B-Chat and Qwen2.5-14B-Instruct demonstrate faster convergence on ALFWorld, typically stabilizing their performance within five trials. 

\section{Conclusion}
This study proposes GA-Rollback, a novel agent framework designed to mitigate the long-standing issue of error propagation, while enhancing the credibility of the reasoning process. GA-Rollback incorporates an independent assistant to support the generator in decision-making and utilizes rollback operations to eliminate potentially erroneous actions. Additionally, we implement probability-based feedback evaluation to improve the assistant’s credibility and introduce Wait-Info strategy for embodied tasks. Extensive experiments across three benchmarks demonstrate that GA-Rollback outperforms several strong baselines and exhibits stronger robustness.
Subsequent results validate the efficacy of each part within the framework. 
Moreover, our findings indicate that GA-Rollback can integrate seamlessly with other methods, underscoring its potential as a plug-and-play module for future applications.

\section*{Limitations}
Despite the improvements achieved by our method, it is important to acknowledge several limitations: 
\begin{enumerate}[leftmargin=*, topsep=1pt, itemsep=1pt, parsep=1pt, label=(\arabic*)]
\item Even with the implementation of feedback evaluation and Wait-Info strategy, GA-Rollback has not yet achieved optimal performance on ALFWorld. This indicates that embodied tasks with longer trajectories pose significant challenges for error detection.
\item The threshold $\theta$ for feedback filtering and the number of Wait-Info steps need to be determined through empirical experiments. Future work could focus on dynamically adjusting these parameters within a single trial.
\item In real-world scenarios, certain environments may not support full rollback to previous states, which could limit the applicability of our method. To address this limitation in tasks involving irreversible actions, we recommend incorporating additional validation steps before executing such actions.
\end{enumerate}

\section*{Acknowledgments}
We would like to thank the anonymous reviewers and meta-reviewer for their insightful suggestions. This work is supported in part by the National Natural Science Foundation of China (62276077, 62406091, U23B2055, 62350710797), in part by the Guangdong Basic and Applied Basic Research Foundation under the Grant No. 2024A1515011205, in part by the Shenzhen Science and Technology Program (ZDSYS20230626091203008, KQTD202402910\\2154066), and in part by the Shenzhen College Stability Support Plan (GXWD20220817123150002, GXWD20220811170358002). This work is also supported by Guangdong Major Project of Basic and Applied Basic Research (Grant No. 2023B0303000010), and is sponsored by CCF-Baidu Open Fund (CCF-Baidu 202404).

\label{sec:bibtex}
\bibliography{custom}

\appendix
\section{Task Details}
\label{app:task_details}
\noindent \textbf{Game of 24} \hspace{0.5em} Game of 24 \cite{tot} is a mathematical reasoning challenge that tests the ability of agents to solve arithmetic puzzles. The task involves using four given numbers and basic arithmetic operations (\opadd, \opsub, \opmul, \opdiv) to obtain a target value of 24. The agent must generate intermediate equations as part of the problem-solving process, and the environment provides feedback on whether the final equation correctly equals 24. In our work, we make subtle modifications to the original environment, enabling it to provide corresponding observations for each action taken by the agent.

\vspace{6pt}
\noindent \textbf{ALFWorld} \hspace{0.5em} ALFWorld \cite{alfworld} presents a series of household tasks that challenge agents to navigate through rooms and apply commonsense reasoning to accomplish tasks, like ``put two soapbar in garbagecan''. The system evaluates the agent's performance by determining if the task is successfully completed within a specified number of steps. The ALFWorld dataset includes both seen and unseen evaluation sets: the seen set evaluates the agent's ability to generalize within the same distribution, while the unseen set, featuring new task scenarios, tests the agent's capacity for out-of-distribution generalization.

\begin{figure*}[t]
\centering
    \begin{minipage}[b]{0.53\linewidth}
        \centering
        \raisebox{-0.5\height}{\scalebox{0.83}{
        \begin{tabular}{lcccc}
        \hline
        \textbf{Method} & \textbf{Avg. Token (K)} & \textbf{Avg. Time (s)} \\
        \hline
        Act-only                & 5.4 & 21.9  \\
        CoT                     & 9.7 & 35.6  \\
        ReAct                   & 9.6 & 30.1  \\
        Reflexion               & 7.2 & 32.2  \\
        ReAct + Reflexion       & 17.1 & 61.9 \\
        \textbf{GA-Rollback}             & 34.1 & 115.6 \\
        \textbf{GA-Rollback} + ReAct     & 40.8 & 162.0 \\
        \textbf{GA-Rollback} + Reflexion & 49.7 & 172.7 \\
        \hline
        \end{tabular}}}
        \vspace{6pt}
        \captionof{table}{Comparison of token consumption and runtime across different methods on Webshop.}
        \label{tab:consumption}
    \end{minipage}
    \hfill  
    \begin{minipage}[b]{0.45\linewidth}
        \centering
        \includegraphics[width=0.9\linewidth]{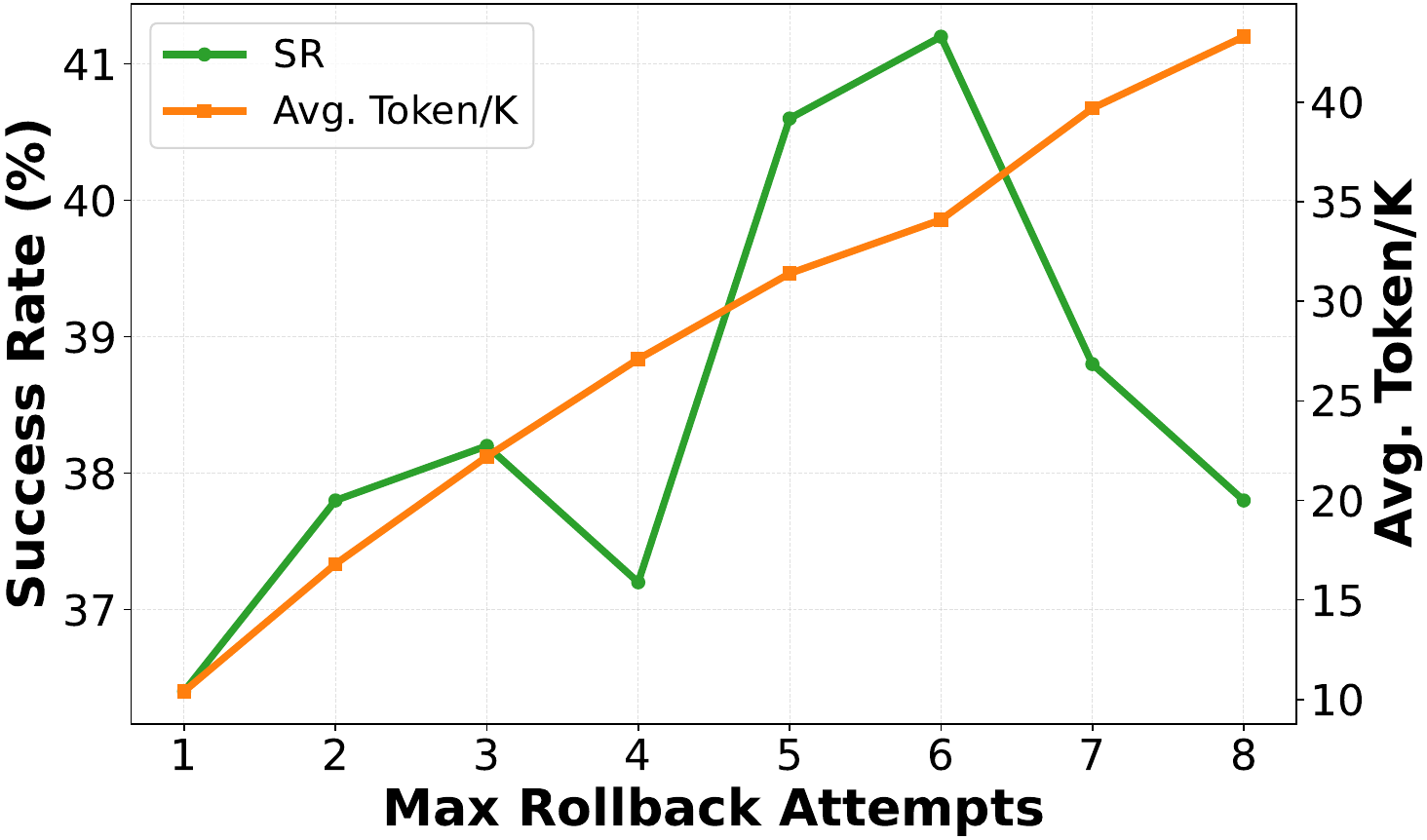}
        \captionof{figure}{Impact of maximum rollback attempts on success rate and token consumption.}
        \label{fig:tradeoff}
    \end{minipage}
\end{figure*}

\vspace{6pt}
\noindent \textbf{Webshop} \hspace{0.5em} Webshop \cite{webshop} is an online simulation environment designed to replicate e-commerce interactions. It features a virtual website containing 1.8 million real-world products, each annotated with unique labels and attributes. Within this environment, agents can perform actions such as ``search'' or ``click'' to navigate and select products that align with given instructions. When the agent chooses the ``buy'' option, the system computes a final reward based on how well the product's attributes and price match the specified criteria. Concretely, the reward is designed as:
\begin{equation}
    r = r_{\mathrm{type}} \cdot \frac{\scriptstyle |U_{\mathrm{att}}\cap Y_{\mathrm{att}}|+|U_{\mathrm{opt}}\cap Y_{\mathrm{opt}}|+\mathbf{1}[y_{\mathrm{price}}\leq u_{\mathrm{price}}]}{\scriptstyle |U_{\mathrm{att}}|+|U_{\mathrm{opt}}|+1},
\end{equation}
where the type reward $r_{\mathrm{type}}=\mathrm{TextMatch}(\overline{y}, \overline{y}^*)$, $U_{\mathrm{att}}$ and $U_{\mathrm{opt}}$ denote the attributes and options of the target product, $Y_{\mathrm{att}}$ and $Y_{\mathrm{opt}}$ represent the attributes and options of the product selected by the agent. 

\section{Additional Results}

\subsection{Consumption Analysis}

Since rollback operations increase communication overhead and time complexity, we further compare the token consumption and runtime of GA-Rollback with other methods using Qwen2.5-14B-Instruct. As shown in Table \ref{tab:consumption}, our GA-Rollback consumes approximately 5–6× more tokens and runtime than Act-only, but achieves significantly higher success rates. To balance success rate and computational cost, we set the maximum rollback attempts per task at 6. Figure \ref{fig:tradeoff} demonstrates that this configuration achieves the optimal trade-off between system performance and resource efficiency, maintaining high success rate while keeping computational cost within reasonable limits.

\subsection{Performance on GPT-4o}
To further validate the generalizability of GA-Rollback, we conduct experiments on the API-based model GPT-4o. We sample 200 instances from the Webshop benchmark to evaluate our approach and baseline methods. Table \ref{tab:gpt4o} shows the effectiveness of our approach.

\begin{table}[ph]
    \centering
    \scalebox{0.83}{
    \begin{tabular}{lw{c}{4.2em}w{c}{4.2em}}
        \toprule
        \multirow{2}{*}{\textbf{Method}} & \multicolumn{2}{c}{\textbf{Webshop}} \\
        \cmidrule(lr){2-3}
        & Reward & SR \\
        \midrule
        Act-only                          & 51.0 & 31.0 \\
        ReAct                             & 36.2 & 21.0 \\
        Reflexion                         & 55.5 & 34.5 \\
        ReAct + Reflexion                 & 40.6 & 25.5 \\
        \textbf{GA-Rollback}              & 47.4 & 32.0 \\
        \textbf{GA-Rollback} + ReAct      & 42.6 & 28.5 \\
        \textbf{GA-Rollback} + Reflexion  & 54.2 & 38.0 \\
        \bottomrule
    \end{tabular}}
    \caption{Performance of different methods on GPT-4o.}
    \label{tab:gpt4o}
\end{table}

\subsection{Results on Different Model Sizes}
\label{app:more_model_results}
To evaluate the generalizability of our proposed method, we conduct experiments across a broader range of model sizes. As presented in Table \ref{tab:diff_scale_model}, the results demonstrate that GA-Rollback consistently brings stable performance gains on both 3B and 32B models. This observation aligns with the trends observed in our main experiments on 8B to 14B models. Moreover, as the model size increases, the overall performance of GA-Rollback improves across all three tasks. These findings confirm the robustness and scalability of our approach in enhancing reasoning accuracy across models of varying sizes.

\begin{table*}[ht]
    \centering
    \scalebox{0.816}{
    \begin{tabular}{cl|w{c}{4.5em}w{c}{4.5em}w{c}{4.5em}w{c}{4.5em}|w{c}{4.5em}}
    \toprule
    \multirow{2}{*}{\textbf{Model}} & \multicolumn{1}{c|}{\multirow{2}{*}{\textbf{Method}}} 
    & \textbf{Game of 24} 
    & \textbf{ALFWorld} 
    & \multicolumn{2}{c|}{\textbf{Webshop}} 
    & \multirow{2}{*}{\textbf{Avg. SR}} \\
    & & SR & SR & Reward & SR \\
    \midrule
    \multirow{8}{*}{\textit{Qwen2.5-3B-Instruct}} & Act-only & 1.0 & 26.1 & 59.9 & 26.4 & 17.8 \\
    & CoT & 0.0 & 37.3 & 62.2 & 30.6 & 22.6 \\
    & ReAct & 1.6 & 44.0 & 39.4 & 17.2 & 20.9 \\
    & Reflexion & 5.6 & 53.7 & 62.1 & 28.0 & \underline{29.1} \\
    & ReAct + Reflexion & \underline{5.8} & \textbf{59.0} & 41.7 & 14.6 & 26.5 \\
    \cmidrule(lr){2-7}
    & \textbf{GA-Rollback} & 5.4 & 41.8 & \underline{64.4} & \underline{34.6} & 27.3 \\
    & \textbf{GA-Rollback} + ReAct & 3.4 & 56.7 & 57.8 & 25.6 & 28.6 \\
    & \textbf{GA-Rollback} + Reflexion & \textbf{6.4} & \underline{57.5} & \textbf{67.4} & \textbf{39.6} & \textbf{34.5} \\
    \midrule
    \midrule
    \multirow{8}{*}{\textit{Qwen2.5-32B-Instruct}} & Act-only & 11.0 & 83.6 & 60.4 & 32.4 & 42.3 \\
    & CoT & 22.8 & 85.1 & 65.5 & 37.8 & 48.6 \\
    & ReAct & 15.8 & 76.9 & 41.3 & 25.4 & 39.4 \\
    & Reflexion & 22.4 & \underline{92.5} & 63.7 & 34.8 & \underline{49.9} \\
    & ReAct + Reflexion & \underline{25.4} & 90.3 & 44.2 & 24.4 & 46.7 \\
    \cmidrule(lr){2-7}
    & \textbf{GA-Rollback} & 20.8 & 88.1 & \underline{66.8} & \underline{38.2} & 49.0 \\
    & \textbf{GA-Rollback} + ReAct & 23.2 & 76.1 & 42.4 & 24.8 & 41.4 \\
    & \textbf{GA-Rollback} + Reflexion & \textbf{33.0} & \textbf{94.8} & \textbf{67.1} & \textbf{43.2} & \textbf{57.0} \\
    \bottomrule
    \end{tabular}
    }
    \caption{The best results of each model are marked in \textbf{bold} and the second-best results are marked with \underline{underline}.}
    \label{tab:diff_scale_model}
\end{table*}

\section{More Cases}
\label{app:cases}
Here, we present more examples to demonstrate the assistant's analysis process on Game of 24 and ALFWorld. The examples are illustrated in Figure \ref{fig:game24_case} and \ref{fig:alf_case}, respectively.

In the \textbf{Game of 24} example, the assistant conducts a thorough analysis and recognizes that the current approach would not yield the desired result of 24. Therefore, it decides to modify the initial step and develops a comprehensive strategy to guide the generator.

In the \textbf{ALFWorld} scenario, although the generator locates both ``\textit{alarmclock 1}'' and ``\textit{desklamp 1}'', it forgets to pick up ``\textit{alarmclock 1}''. Through a step-by-step inspection, the assistant identifies this oversight and triggers a rollback operation, effectively preventing subsequent invalid actions. These two examples demonstrate the assistant's critical role in identifying and rectifying errors, which contributes significantly to optimizing the overall trajectory of task execution.

Furthermore, in Figure \ref{fig:conflict_case}, we present an example of a conflict between ReAct and GA-Rollback. In this case, the generator concludes that none of the retrieved items match the requirement ``\textit{redwood color}'', while the assistant maintains that a check of at least one item is necessary before making such a determination. This divergence in judgment leads to redundant rollback operations, which could potentially disrupt subsequent actions.

\begin{figure*}[htp]
    \centering
    \begin{tcolorbox}[
        colback=mygray, 
        colframe=darkgray, 
        colbacktitle=darkgray, 
        coltitle=white, 
        arc=4pt, 
        title=Case Study for Game of 24, 
        width=\linewidth,
    ]
    \begin{tabular}{p{0.97\linewidth}}

\textcolor{darkgreen}{\textbf{Observation 0:}} \\
\# Here is the task: \\
Input: 1 2 10 13 \\
\textcolor{darkorange}{\textbf{Action 1:}} 13 - 10 = 3 \\
\textcolor{darkgreen}{\textbf{Observation 1:}} numbers left: 1 2 3 \\
\textcolor{darkorange}{\textbf{Action 2:}} 3 * 2 = 6 \\
\textcolor{darkgreen}{\textbf{Observation 2:}} numbers left: 1 6 \\
\addlinespace
\hline
\addlinespace
\textcolor{myblue}{\textbf{Analysis Process by Assistant:}} \\
\textcolor{darkred}{\textbf{Let's analyze the current trajectory step by step:}}\\
- Action 1: 13 - 10 = 3\\
\hspace{0.6em}  - This operation is valid and reduces the numbers to 1, 2, and 3.\\
- Action 2: 3 * 2 = 6\\
\hspace{0.6em}  - This operation is also valid and further reduces the numbers to 1 and 6.\\
However, \textcolor{darkred}{\textbf{the trajectory ends here without reaching the target value of 24.}} We need to continue the steps to achieve the final result.\\
\vspace{1pt}
\#\#\# Next Steps:\\
To reach 24, we need to use the remaining numbers (1 and 6) in a way that results in 24. One possible continuation could be:\\
- Action 3: 6 * 1 = 6 (This doesn't help, so we need another approach)\\
- Alternatively, we can try:\\
\hspace{0.6em}  - Action 3: 6 + 1 = 7 (Still not helping, so we need another approach)\\
\textcolor{darkred}{\textbf{Since the current trajectory does not lead to 24, we need to re-evaluate the initial steps to find a more effective path.}}\\
\vspace{1pt}
\#\#\# \textcolor{darkred}{\textbf{Correct Approach:}}\\
Given the numbers 1, 2, 10, and 13, one effective way to reach 24 is:\\
- Action 1: 13 - 1 = 12\\
- Action 2: 12 + 2 = 14\\
- Action 3: 14 + 10 = 24\\
Thus, the correct trajectory should be:\\
- Action 1: 13 - 1 = 12\\
- Action 2: 12 + 2 = 14\\
- Action 3: 14 + 10 = 24\\
- Action 4: answer: (13 - 1) + 2 + 10 = 24\\
\vspace{1pt}
\#\#\# Conclusion\\
\textcolor{darkred}{\textbf{Error Location: Action 1}}\\
Explanation: The initial subtraction in Action 1 (13 - 10 = 3) led to a dead end because it made it difficult to reach 24 with the remaining numbers. A better approach would be to subtract 1 from 13 to get 12, then add 2 and 10 to reach 24. \textcolor{darkred}{\textbf{The correct expression should be: "answer: (13 - 1) + 2 + 10 = 24".}} This ensures that all numbers are used correctly to achieve the target value of 24.\\
\addlinespace
\hline
\addlinespace
\textbf{-- Rollback Happened --} \\
\textcolor{darkorange}{\textbf{Action 1:}} 13 - 1 = 12
    \end{tabular}
    \end{tcolorbox}
    \caption{Case study for Game of 24.}
    \label{fig:game24_case}
\end{figure*}

\begin{figure*}[ht]
    \centering
    \begin{tcolorbox}[
        colback=mygray, 
        colframe=darkgray, 
        colbacktitle=darkgray, 
        coltitle=white, 
        arc=4pt, 
        title=Case Study for ALFWorld, 
        width=\linewidth
    ]
    \begin{tabular}{p{0.97\linewidth}}
\textcolor{darkgreen}{\textbf{Observation 0:}}\\
You are in the middle of a room. Looking quickly around you, you see a bed 1, a desk 2, a desk 1, a drawer 6, a drawer 5, a drawer 4, a drawer 3, a drawer 2, a drawer 1, a garbagecan 1, a laundryhamper 1, a safe 1, a shelf 6, a shelf 5, a shelf 4, a shelf 3, a shelf 2, and a shelf 1.\\
\textcolor{darkgreen}{\textbf{Your task is to:}} examine the alarmclock with the desklamp.\\
\textcolor{darkorange}{\textbf{Action 1:}} go to desk 1\\
\textcolor{darkgreen}{\textbf{Observation 1:}} On the desk 1, you see a bowl 1, a desklamp 1, and a laptop 2.\\
\textcolor{darkorange}{\textbf{Action 2:}} use desklamp 1\\
\textcolor{darkgreen}{\textbf{Observation 2:}} You turn on the desklamp 1.\\
\textcolor{darkorange}{\textbf{Action 3:}} go to desk 2\\
\textcolor{darkgreen}{\textbf{Observation 3:}} On the desk 2, you see a alarmclock 1, a keychain 2, and a mug 1.\\
\textcolor{darkorange}{\textbf{Action 4:}} examine alarmclock 1 with desklamp 1\\
\textcolor{darkgreen}{\textbf{Observation 4:}} Nothing happens.\\
\textcolor{darkorange}{\textbf{Action 5:}} go to desk 1\\
\textcolor{darkgreen}{\textbf{Observation 5:}} On the desk 1, you see a bowl 1, a desklamp 1, and a laptop 2.\\
\textcolor{darkorange}{\textbf{Action 6:}} take desklamp 1 from desk 1\\
\textcolor{darkgreen}{\textbf{Observation 6:}} Nothing happens.\\
\textcolor{darkorange}{\textbf{Action 7}}: use desklamp 1\\
\textcolor{darkgreen}{\textbf{Observation 7}}: You turn on the desklamp 1.\\
\addlinespace
\hline
\addlinespace
\textcolor{myblue}{\textbf{Analysis Process by Assistant:}} \\
\textcolor{darkred}{\textbf{Let's analyze it step by step from back to front.}}\\
\#\#\# Step-by-Step Analysis\\
1. Action 7: use desklamp 1\\
\hspace{0.6em}   - Observation 7: You turn on the desklamp 1.\\
\hspace{0.6em}   - This action is valid, but it does not help in examining the alarmclock.\\
2. Action 6: take desklamp 1 from desk 1\\
\hspace{0.6em}   - Observation 6: Nothing happens.\\
\hspace{0.6em}   - This indicates that the desklamp 1 cannot be taken from the desk 1. This could be because the desklamp is fixed or not designed to be moved.\\
3. Action 5: go to desk 1\\
\hspace{0.6em}   - Observation 5: On the desk 1, you see a bowl 1, a desklamp 1, and a laptop 2.\\
\hspace{0.6em}   - This confirms that the desklamp 1 is still on the desk 1.\\
4. Action 4: examine alarmclock 1 with desklamp 1\\
\hspace{0.6em}   - Observation 4: Nothing happens.\\
\hspace{0.6em}   - \textcolor{darkred}{\textbf{This indicates that the action is invalid. The desklamp 1 must be used in a different way or the alarmclock 1 must be brought to the desklamp 1.}}     \textbf{......} \\
\vspace{1pt}
\#\#\# Conclusion\\
\textcolor{darkred}{\textbf{Error Location: Action 4}}\\
Explanation: The desklamp 1 cannot be used directly to examine the alarmclock 1 while it is on desk 2. \textcolor{darkred}{\textbf{Instead, I should take the alarmclock 1 from desk 2 and bring it to desk 1 where the desklamp 1 is located.}} After that, I can use the desklamp 1 to examine the alarmclock 1.\\
\addlinespace
\hline
\addlinespace
\textbf{-- Rollback Happened --} \\
\textcolor{darkorange}{\textbf{Action 4:}} take alarmclock 1 from desk 2
    \end{tabular}
    \end{tcolorbox}
    \caption{Case study for ALFWorld.}
    \label{fig:alf_case}
\end{figure*}

\begin{figure*}[ht]
    \centering
    \begin{tcolorbox}[
        colback=mygray, 
        colframe=darkgray, 
        colbacktitle=darkgray, 
        coltitle=white, 
        arc=4pt, 
        title=A Conflict Example of ReAct and GA-Rollback, 
        width=\linewidth
    ]
    \begin{tabular}{p{0.97\linewidth}}
\textcolor{darkgreen}{\textbf{Observation 0:}} \\
i'm looking for a queen size bedspread set in the color redwood, and price lower than 60.00 dollars \\
{\textbf{[Search]}} \\
\textcolor{darkorange}{\textbf{Action 1:}} search{[queen size bedspread set redwood]}\\
\textcolor{darkgreen}{\textbf{Observation 1:}} \\
{\textbf{[Back to Search]}} \\
Page 1 (Total results: 50) \\
{\textbf{[Next >]}} \\
{\textbf{[B07B4KXQZV]}} \\
Lodge Bedspread Full/Queen Size Quilt with 2 Shams. Cabin 3-Piece Reversible All Season Quilt Set. Rustic Quilt Coverlet Bed Set. Stonehurst Collection. \\
\$44.99 \\
{\textbf{[B07PGQ3JYB]}} \\
Lodge Bedspread Full/Queen Size Quilt with 2 Shams. Cabin 3-Piece Reversible All Season Quilt Set. Rustic Quilt Coverlet Bed Set. Wilderness Collection (Moose - Grey) \\
\$44.99 \\
{\textbf{[B07WZBXSK3]}} \\
Travan 3-Piece Queen Quilt Set Quilted Bedspread Oversized Microfiber Lightweight Bedspread Coverlet Set with Shams Bedding Set for All Season, Grey, Queen \\
\$64.99 \\
\textcolor{darkorange}{\textbf{Action 2:}} \textcolor{darkred}{\textbf{think[None of the products returned in the search results match the color redwood. I should consider searching for similar colors or broader terms.]}}\\
\textcolor{darkgreen}{\textbf{Observation 2:}} OK.\\
\addlinespace
\hline
\addlinespace
\textcolor{myblue}{\textbf{Analysis Process by Assistant:}} \\
Let's analyze it step by step from back to front.\\
- Observation 2 shows that the agent is thinking about checking for similar colors since there are no items matching the exact color redwood. However, the agent hasn't taken any action yet.\\
- Observation 1 shows that there are three options available, but none of them are in the color redwood. The agent should proceed to check for similar colors as mentioned in Observation 2.\\
[10pt]
\#\#\# Conclusion\\
\textcolor{darkred}{\textbf{Error Location: Action 2}}\\
Explanation: \textcolor{darkred}{\textbf{The agent thought about checking for similar colors but did not take any action. The agent should click on one of the products and check if the color is similar to redwood or modify the search query to include similar colors.}} For example, the agent could click on \textbf{[B07B4KXQZV]} and check if the color is similar to redwood or search for "queen size bedspread set redwood similar". 
    \end{tabular}
    \end{tcolorbox}
    \caption{A conflict example of ReAct and GA-Rollback.}
    \label{fig:conflict_case}
\end{figure*}

\onecolumn

\newpage
\section{Prompt for Assistant}
The instruction prompts specifically designed for the assistant in Game of 24, ALFWorld, and Webshop are presented in Figures \ref{fig:prompt_game}, \ref{fig:prompt_alf}, and \ref{fig:prompt_web}, respectively.

\begin{figure}[hp]
    \centering
    \begin{tcolorbox}[
        colback=mygray, 
        colframe=darkgray, 
        colbacktitle=darkgray, 
        coltitle=white, 
        arc=4pt, 
        title=Instruction Prompt for the Assistant in Game of 24, 
        width=\linewidth
    ]
\# \textbf{Task Description}\\
You will analyze trajectories for potential errors and provide analysis in a standardized format.\\
\\
\# \textbf{Analysis Rules}\\
1. Focus on Error Detection: Analyze each step in the trajectory to identify errors, regardless of the current task status.\\
\\
2. Analysis Method:\\
    - Analyze the given trajectory step by step\\
    - Identify where the trajectory first deviated from correct behavior\\
\\
3. Required Checks:\\
    - If the number obtained after three operations is not 24, the calculation process should be changed\\
    - If the final result is 24, then it should be displayed as a complete expression using the four numbers in the input, such as "answer: (6 - 4) * (4 + 8) = 24"\\
\\
4. Output Must Include:\\
    - ** Error Location **: Specific step where error occurred\\
    - ** Explanation **: Error explanation and correction method\\
\\
\# \textbf{Examples}\\
\{Analysis Examples\}\\
\\
\# \textbf{Current Task}\\
\{Trajectory to be analyzed\}
    \end{tcolorbox}
    \caption{Instruction prompt for the assistant in Game of 24.}
    \label{fig:prompt_game}
\end{figure}

\begin{center}
    \begin{tcolorbox}[
        colback=mygray, 
        colframe=darkgray, 
        colbacktitle=darkgray, 
        coltitle=white, 
        arc=4pt, 
        title=Instruction Prompt for the Assistant in ALFWorld, 
        width=\linewidth,
        breakable,
    ]
\# \textbf{Task Description}\\
You will analyze trajectories for potential errors and provide analysis in a standardized format.\\
\\
\# \textbf{Analysis Rules}\\
1. Focus on Error Detection: Analyze each step in the trajectory to identify errors, regardless of the current task status.\\
\\
2. Analysis Method:\\
    - Start from the final outcome\\
    - Work backwards through each action\\
    - Identify where the trajectory first deviated from correct behavior\\
\\
3. Required Checks:\\
    - Verify that you have reached the correct location before picking up objects\\
    - Confirm destination exists in the environment before navigation\\
    - If the action 'go to [location]' results in 'Nothing happens', it means you are already at that location or the location doesn't exist in the environment\\
\\
4. Output Must Include:\\
    - ** Error Location **: Specific step where error occurred\\
    - ** Explanation **: Error explanation and correction method\\
\\
\# \textbf{Examples}\\
\{Analysis Examples\}\\
\\
\# \textbf{Current Task}\\
\{Trajectory to be analyzed\}
    \end{tcolorbox}
    \captionof{figure}{Instruction prompt for the assistant in ALFWorld.}
    \label{fig:prompt_alf}
\end{center}

\begin{figure}[hp]
    \centering
    \begin{tcolorbox}[
        colback=mygray, 
        colframe=darkgray, 
        colbacktitle=darkgray, 
        coltitle=white, 
        arc=4pt, 
        title=Instruction Prompt for the Assistant in Webshop, 
        width=\linewidth
    ]
\# \textbf{Task Description}\\
You will analyze trajectories for potential errors and provide analysis in a standardized format.\\
\\
\# \textbf{Analysis Rules}\\
1. Focus on Error Detection: Analyze each step in the trajectory to identify errors, regardless of the current task status.\\
\\
2. Analysis Method:\\
    - Start from the final outcome\\
    - Work backwards through each action\\
    - Identify where the trajectory first deviated from correct behavior\\
\\
3. Required Checks:\\
    - Verify all required product attributes are selected\\
    - Confirm specifications match the user's requirements\\
    - Ensure necessary clicks/selections are made before purchase\\
\\
4. Output Must Include:\\
    - ** Error Location **: Specific step where error occurred\\
    - ** Explanation **: Error explanation and correction method\\
\\
\# \textbf{Examples}\\
\{Analysis Examples\}\\
\\
\# \textbf{Current Task}\\
\{Trajectory to be analyzed\}
    \end{tcolorbox}
    \caption{Instruction prompt for the assistant in Webshop.}
    \label{fig:prompt_web}
\end{figure}

\end{document}